\documentclass[journal]{IEEEtran}

\usepackage{gensymb}
\usepackage{graphicx}
\usepackage{amsmath}
\DeclareMathOperator{\sign}{sign}
\usepackage{caption}
\usepackage{subcaption}
\usepackage{soul}
\usepackage[dvipsnames]{xcolor}
\newcommand{\added}[1]{#1}
\newcommand{\removed}[1]{}
\newcommand{\rev}[1]{#1}
\newcommand{\rem}[1]{}

\usepackage{tabularx}
\definecolor{Gray}{gray}{0.90}
\usepackage{colortbl}

\usepackage[most]{tcolorbox}
\usepackage{multicol}
\usepackage{siunitx}  
\usepackage{bm}
\usepackage{siunitx}
\usepackage{hyperref}

\usepackage{textcomp}
\usepackage[backend=biber,      %
    style=ieee,
    bibstyle=ieee,              %
    mincitenames=1,
    maxcitenames=2,
    sortcites=true,   
    giveninits=true,            %
    backref=false
]{biblatex}
\title{Learning Fast and Precise Pixel-to-Torque Control} %

\begin{document}

\author{Steffen Bleher$^1$, Steve Heim$^{1,2}$, Sebastian Trimpe$^{1,3}$\\
$^1$ Intelligent Control Systems Group, Max Planck Institute for Intelligent Systems, Stuttgart, Germany; \\
$^2$ Biomimetic Robotics Lab, Massachusetts Institute of Technology, Cambridge, USA; \\
$^3$ Institute for Data Science in Mechanical Engineering, RWTH Aachen University, Aachen, Germany}
\maketitle

\section{Introduction}
In the field, robots often need to \removed{reason and plan}\added{operate} in unknown and unstructured environments, where accurate sensing and state estimation (SE) becomes a major challenge.
Cameras have been used to great success in mapping and planning in such environments~\cite{ostafew2016learning}, as well as complex but quasi-static tasks such as grasping~\cite{graspingLevine}, but are rarely integrated into the control loop for unstable systems.
Learning pixel-to-torque control promises to allow robots to flexibly handle a wider variety of tasks.
\rem{However, especially for unstable systems that require precise and high bandwidth control, learning pixel-to-torque control still poses a significant challenge, and best practices have not yet been established.}
\rev{Although they do not present additional theoretical obstacles, learning pixel-to-torque control for unstable systems that that require precise and high bandwidth control still poses a significant practical challenge, and best practices have not yet been established.}
Part of the reason is that many of the most auspicious tools, such as deep neural networks (DNN), are opaque: the cause for success on one system is difficult to interpret and generalize.
\par
\begin{figure}[!t]
\centering
\includegraphics[width=\linewidth]{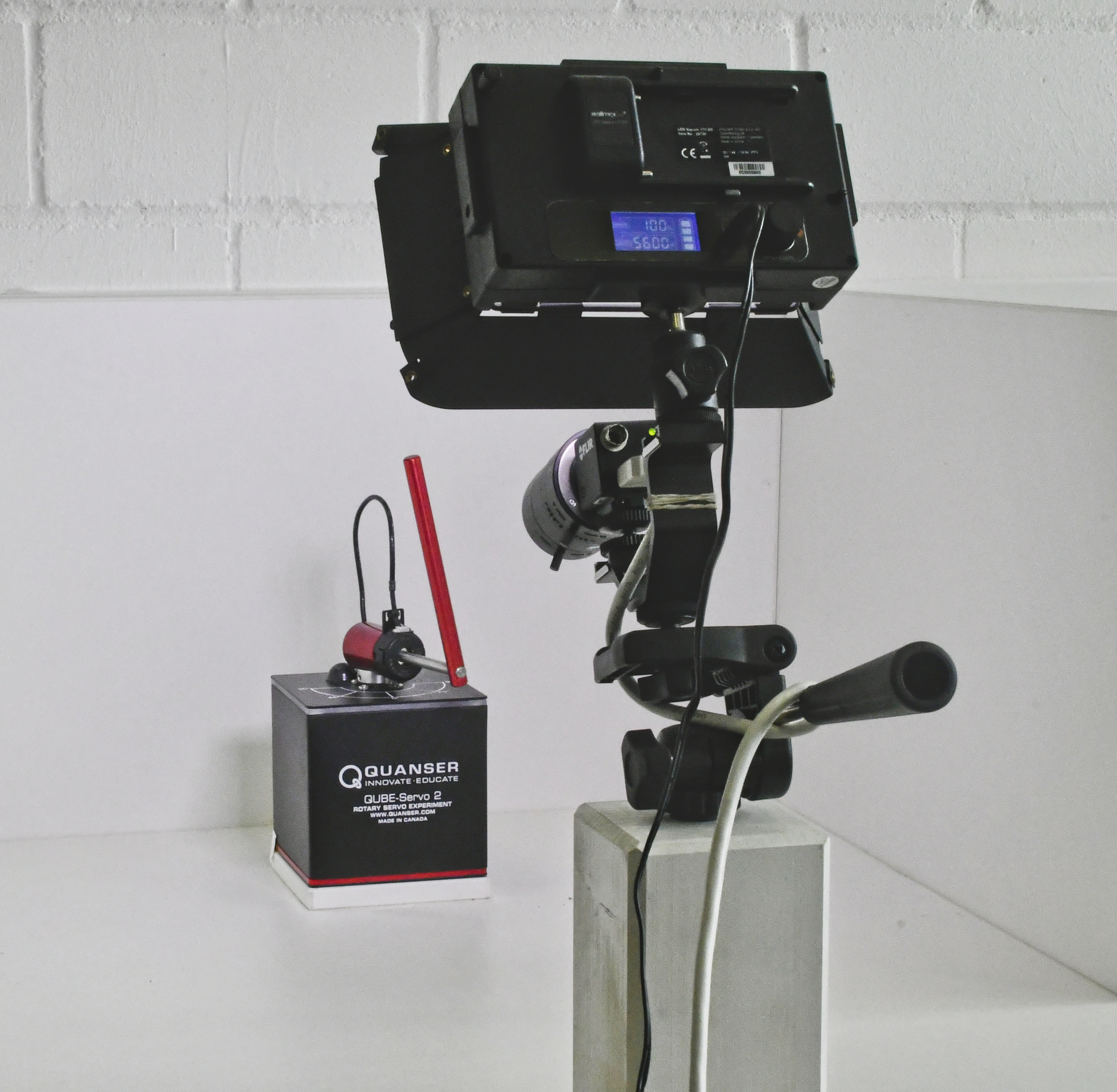}
\caption{The vision-based Furuta pendulum: a platform for reproducible research on learning fast and precise pixel-to-torque control.}
\label{pendulum-introduction}
\end{figure}

The machine learning community has alleviated this problem by establishing standard data sets and standardized simulation environments that allow different approaches to be easily benchmarked against each other.
This trend is not well established in the robotics community, as there are many more hurdles to reproduce a system in hardware than purely in simulation.
\rem{To address these issues,}\rev{To help drive reproducible research on the practical aspects of learning pixel-to-torque control,} we propose a platform that can flexibly represent the entire process, from lab to deployment, for learning pixel-to-torque control on a robot with fast, unstable dynamics: the vision-based Furuta pendulum.
The platform, shown in Figure~\ref{pendulum-introduction} and detailed in ``Reproducible Platform'', can be reproduced \added{with either off-the-shelf or custom-built hardware.} \removed{for under 10'000 € with off-the-shelf hardware.}
We expect that this platform will allow researchers to quickly and systematically test different approaches, as well as reproduce and benchmark case studies from other labs.
\par
We also present a first case study on this system using DNNs which, to the best of our knowledge, is the first demonstration of learning pixel-to-torque control on an unstable system with update rates faster than $\SI{100}{\hertz}$.
\rem{A video synopsis is included in the supplementary material and will be made publicly available upon publication.}
\rev{A video synopsis can be found online at \url{https://youtu.be/S2llScfG-8E}, and in the supplementary material.}

\added{\section{Related Work}}
DNNs combined with RL have had tremendous success recently in a variety of robotics applications~\cite{lee2020terrainLocomotion, openai2019solving, finn2016spatialAutoencoder, graspingLevine}, though many open challenges remain~\cite{ibarz2021train}. 
One of the most critical challenges in all these approaches is sample efficiency---or rather, sample inefficiency. In most cases, learning is done in simulation only, which adds the challenge of transferring the learned policy from simulation to the actual hardware.
To make this transfer successfully, a lot of effort is typically put into modeling and system identification~\cite{openai2019solving, xie2020learning}, such that the gap between simulation and reality is `small' in some sense.
For certain dynamics, such as turbulent flows or soft matter~\cite{Roberts2010flapping, rohr2018softlight}, accurate models are unavailable or prohibitive to simulate.
Overcoming the sample-efficiency challenge would not only allow learning to be leveraged on these systems, but also alleviate the reality gap in general: policies can be refined on the real hardware after initial training in a low-fidelity simulation.
Indeed,~\textcite{lee2020terrainLocomotion} point out that, while they needed a high-fidelity model of the robot dynamics to train a DNN policy, they could successfully make the transfer from simulation to reality using only low-fidelity terrain models.
\par
One of the key concepts they leverage is \emph{privileged information}: ground-truth data is often available during training, and can be leveraged to substantially reduce training time.~\textcite{dian2019cheating} coined this term, and use it to improve imitation learning by first training an autonomous driving policy with a birds-eye view of the environment, then using its evaluations as training examples for the final policy, which only has access to a regular car-mounted camera as input.
~\textcite{lee2020terrainLocomotion} use the same concept to infer terrain properties from a history of proprioceptive data and thus avoid the need for external sensing entirely.
In both these studies, learning is done in simulation, and privileged information can be directly accessed from the simulator.
Privileged information is also available when learning directly in hardware,~\removed{if, as is often the case,}\added{especially if} it takes place in a controlled lab setting.
\added{For example,~\textcite{Srinivasan2020AutoRaceSE} learn accurate SE for a racing car from only IMU and wheel encoder readings.
Training targets are generated with a mixed Kalman filter that has access to two additional velocity sensors, which are very accurate but also expensive.
Previously, }~\textcite{levine2016end} used this concept to learn to estimate the position of a target object from images, using supervised learning.
During this phase, the object is placed in the robot's gripper, so the robot can directly estimate its position relative to the camera through joint-position measurements and forward kinematics.
\par
\added{\textcite{levine2016end} also report significantly better performance with full end-to-end learning, that is, training a single DNN for both the state estimator and controller. However, separating SE and control also has benefits, such as improved sample efficiency and more targeted development.
For example, ~\textcite{Srinivasan2020AutoRaceSE} rely on existing methods for perception, mapping, and control, and focus on learning a convolutional DNN specifically to estimate velocities, which can be challenging for a Kalman filter during aggressive maneuvers.
~\textcite{Hoeller21StateRep} implement a full, highly modular learning pipeline, which separately tackles state-representation\footnote{State representation only differs from state estimation in that it includes relevant state of the environment, such as moving obstacles.} and motion planning. This pipeline is trained to high performance with remarkable sample efficiency, requiring on the order of 70'000 depth-images and 17 hours worth of trajectories, using a mixture of simulated and real-world data.}
\par
Despite recent successes in learning vision-based controllers for grasping~\cite{graspingLevine,zeng2019tossingbot, ha2021flingbot}, learning pixel-to-torque control remains elusive, especially for fast, unstable systems.
\removed{An important difference is that tasks such as grasping, although challenging in their own respects, do not require learning controllers that act directly on the system dynamics: instead, the learned policies can output desired kinematics and rely on a conventional low-level controller to track these.}
\added{An important difference is that for tasks such as grasping, a conventional low-level controller can be relied on to stabilize the dynamics. Instead, the challenge is to generate appropriate desired kinematics such as grasping positions~\cite{graspingLevine, ha2021flingbot}, or kinematic trajectories, often called primitives~\cite{zeng2019tossingbot, ha2021flingbot}.
In other words, learning is used for \emph{planning}, rather than for \emph{control}.}
\par
Since torque control is usually required when systems have fast and unstable dynamics, high control bandwidth is often a concern when learning pixel-to-torque control. 
This need for fast and precise feedback is a key characteristic of the proposed platform, which distinguishes it from more common platforms for research on vision-based \removed{control and RL}\added{learning}.
\par
\textcite{lambert2019lowLevelQuadrotor} use DNNs to learn a predictive low-level controller of a hovering quadcopter using onboard sensors as input, and manage to obtain stable hovering for several seconds at a time after training on only a few minutes of data.
Despite running a relatively simple DNN architecture on a powerful, offboard GPU, evaluation time is the bottleneck: to obtain sufficiently long prediction horizons requires multiple evaluations of the DNN, which limits their control bandwidth to \SI{50}{\Hz}.
This bottleneck is greatly exasperated when vision is used for feedback since the high dimensions and complexity of vision typically require larger and more sophisticated DNNs, which are consequently slower to evaluate.
For example,~\textcite{mattner2012pendulum} use an auto-encoder architecture to learn pixel-to-torque control for balancing an inverted pendulum with minimal domain knowledge. The entire DNN size is kept manageable in a number of ways, including down-sampling the input image to 40\texttimes30 pixels and limiting the output torque to only three values.
Nonetheless, the control bandwidth is limited to roughly \SI{10}{\Hz}.
Fast evaluation becomes even more challenging for autonomous robots that rely on onboard computation, since more powerful computers add strain to a limited payload and battery supply.
To run vision-based control fully onboard a small quadcopter,~\textcite{kaufmann2018deepRacing} only learn a DNN for part of the control pipeline, which runs at a lower frequency of \SI{10}{\Hz}.
Similar to the grasping examples above, stability is maintained by a conventional controller running at a higher frequency, in this case a minimum-jerk planner.
To learn the full pixel-to-torque control pipeline of fast systems, the trade-off between the learned SE's precision and its evaluation speed takes a central role in designing the learning pipeline, as we will explore in our case study.

\begin{figure*}[htp]
	\begin{tcolorbox}[title=Reproducible Platform]
    \begin{multicols}{2}
        \input{sections/furuta}
    \end{multicols}
\end{tcolorbox}
\end{figure*}
\section{Learning Pipeline}
\begin{figure*}[!t]
\centering
\includegraphics[width=\linewidth]{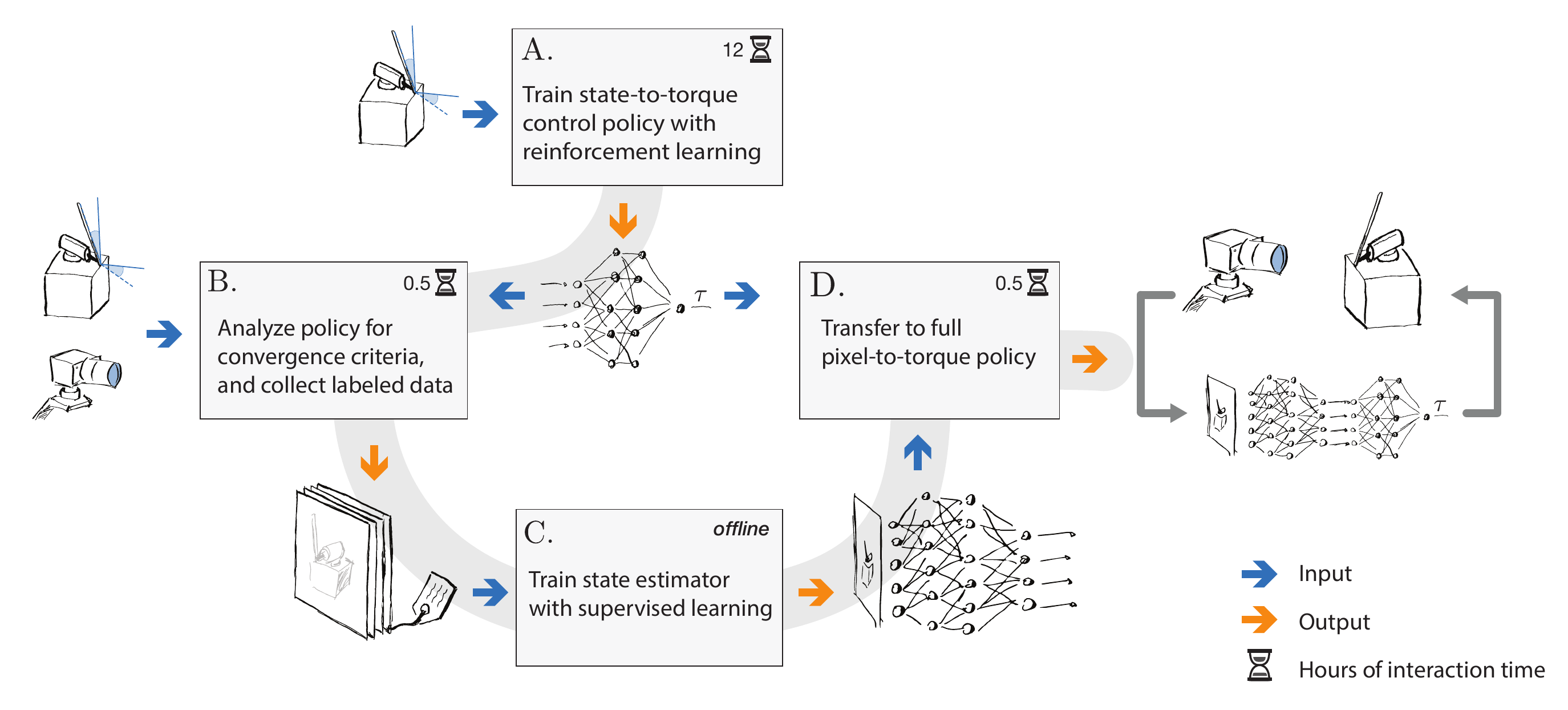}
\caption{%
The four steps of the pipeline can be completed in less than 18 hours, with only 13 hours of hardware interaction time, which is automated largely automated. Note that step \emph{C} is done completely offline and can be accelerated by running the training on a dedicated cluster.
Privileged information is used in steps \emph{A} and \emph{B}.
}
\label{full-process}
\end{figure*}

We found the two central criteria for designing the learning pipeline to be sample efficiency, and simultaneously fulfilling the precision and control bandwidth requirement.
For the Furuta pendulum used in this study, the minimum control frequency translates to a time budget for the entire vision-based control of roughly \SI{8}{\milli\s} (see ``Reproducible Platform'').
We found it essential to split up the learning pipeline into four steps (see Figure \ref{full-process}): in step \emph{A}, online RL of a control policy using privileged information, in step \emph{B}, policy analysis and sample-collection using privileged information, in step \emph{C}, offline supervised learning of the SE, and finally in step \emph{D}, online adaptation learning of the control policy to the SE.\par
We rely on privileged information in the form of rich and accurate state measurements, which are often available in the lab setting via external sensing such as motion capture. We also rely on a means of automating sample collection with a specified distribution.
In the case of the Furuta pendulum, state measurements are readily available from the joint encoders, and samples can be easily gathered using a standard combination of energy-pumping and LQR controllers.
An important benefit of automating sample collection is that it makes it possible to quickly and easily collect new data sets.
As is always the case, development is an iterative process, and speeding up this process is critical yet seldomly discussed in literature~\cite{xie2020learning}.
\par

\subsection{Learning the Control Policy}\label{control-policy}

To focus on a reliable and sample efficient training process, we train a Proximal Policy Optimization (PPO)~\cite{schulman2018Proximal} RL agent using privileged information as input. 
In the case of the Furuta platform, the agent learns to swing up and balance the pole in approximately \SI{12}{\hour} of interaction time, which is equivalent to \SI{8}{\hour} worth of samples gathered for learning and \SI{4}{\hour} for resetting. The entire process is automated and could be run in a single session without any intervention.
\par
To enable the agent to learn this task reliably, it was important to tune the reward function and to adjust hyperparameters based on knowledge about the system.
We use a continuous reward function, which accelerates training by providing a reliable, steady increase in the accumulated reward.
For the Furuta pendulum, we use a quadratic reward penalizing the angle positions of the pendulum with

\begin{equation}
    r_t = \left(1 - \frac{4}{5} \frac{|\alpha_t|}{\SI{180}{\degree}} - \frac{1}{5} \frac{|\theta_t|}{\SI{180}{\degree}} \right)^2.
\end{equation}

We train the agent with a small learning rate and a small clipping factor (compare Table \ref{rl-hyperparameters}), which also helps to reliably increase the reward over training episodes.
Agents with a large learning rate learned the swing-up task more quickly, but were not able to learn to balance the pendulum reliably: 
they were susceptible to `fatal forgetting', or sudden large drops in reward. We surmise this is because balancing requires very precise control inputs, and therefore also a smaller learning rate.
\par

\subsection{Policy Analysis and Data Collection}\label{sub:analysis}

Based on the control policy trained on privileged information, we empirically identify minimum precision requirements by injecting noise on the state until the task can no longer be fulfilled.
This threshold is then used as the convergence criteria for training the SE in step~\ref{sub:estimator}.
For the Furuta pendulum, we add zero-mean Gaussian noise on the angles, and propagate it via finite differences to the angular velocities.
At a sampling frequency of $\SI{120}{\hertz}$, the agent can tolerate noise with a standard deviation $<\SI{1}{\degree}$.
We also noticed that this level of precision is only necessary to balance the pendulum near the equilibrium point; the policy is able to swing up the pendulum even with higher noise.
Based on this observation, we separately collect data for the swing-up portion of the task, and the balancing portion (see ``Reproducible Platform'').
The convergence criteria is then only tested on images relevant for balancing, which we heuristically determined as $|\alpha| < \SI{10}{\degree}$.
As we will see in Section~\ref{sub:estimator}, converging to high precision over the entire state space not only requires more training time, but a larger DNN.
\subsection{Learning precise State Estimation \label{sub:estimator}}
Precise predictions require an unknown minimum network capacity, which makes it difficult to reduce execution time on limited computational resources.
We balance this trade-off with a deliberate choice of the DNN architecture, a biased data set, and data augmentation methods.\par
\begin{figure*}
\centering
\includegraphics[width=\linewidth]{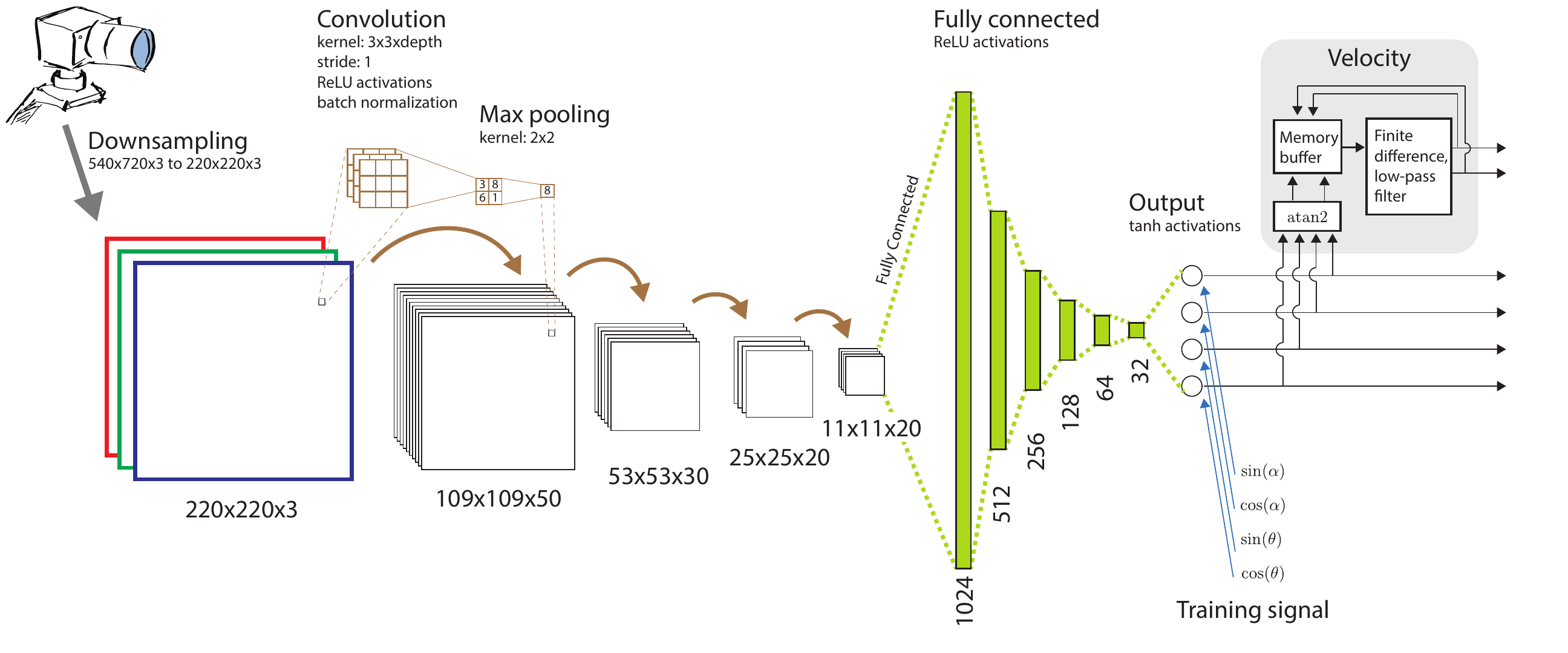}
\caption{The architecture of the SE. Instead of predicting the angles directly, we train the DNN to predict $[\cos\theta, \sin\theta, \cos\alpha, \sin\alpha]^T$. We thereby ensure that samples that are very close to each other in the input space are also close in the output space by avoiding jumps for the angles around $\pm \SI{180}{\degree}$. During training, we apply neuron dropouts with a probability of 0.1 after every max-pooling layer and every fully connected layer to prevent overfitting to the training data set.}
\label{final-architecture}
\end{figure*}

To increase precision, we simplify the learning task and train a DNN using standard convolutional layers to estimate only the pose from a single image.
Velocities are then computed from a buffer of previously estimated positions and velocities via finite differences and a first-order low pass filter (see Figure~\ref{final-architecture}).
This structure reduces the SE's prediction error to roughly a fifth compared to a recurrent neural network architecture similar in size, which we speculate is due to the freed capacity being available for higher accuracy on a simpler task.
Alternatively, velocities could be estimated by using a history of images as input, but again this would significantly increase the network size, which we need to reduce as much as possible.
\par

We also downsample the input image from 540\texttimes720 to 220\texttimes220 pixels, which allows the DNN depth to be increased; we found this was more important for precision than a higher image resolution.
To compensate for the downsampling, we add a very small stride of 1-pixel per step. With a depth of 12 layers, the SE reaches a precision that that is able to distinguish individual pixels.
\par
\begin{figure*}
\centering
\includegraphics[width=\linewidth]{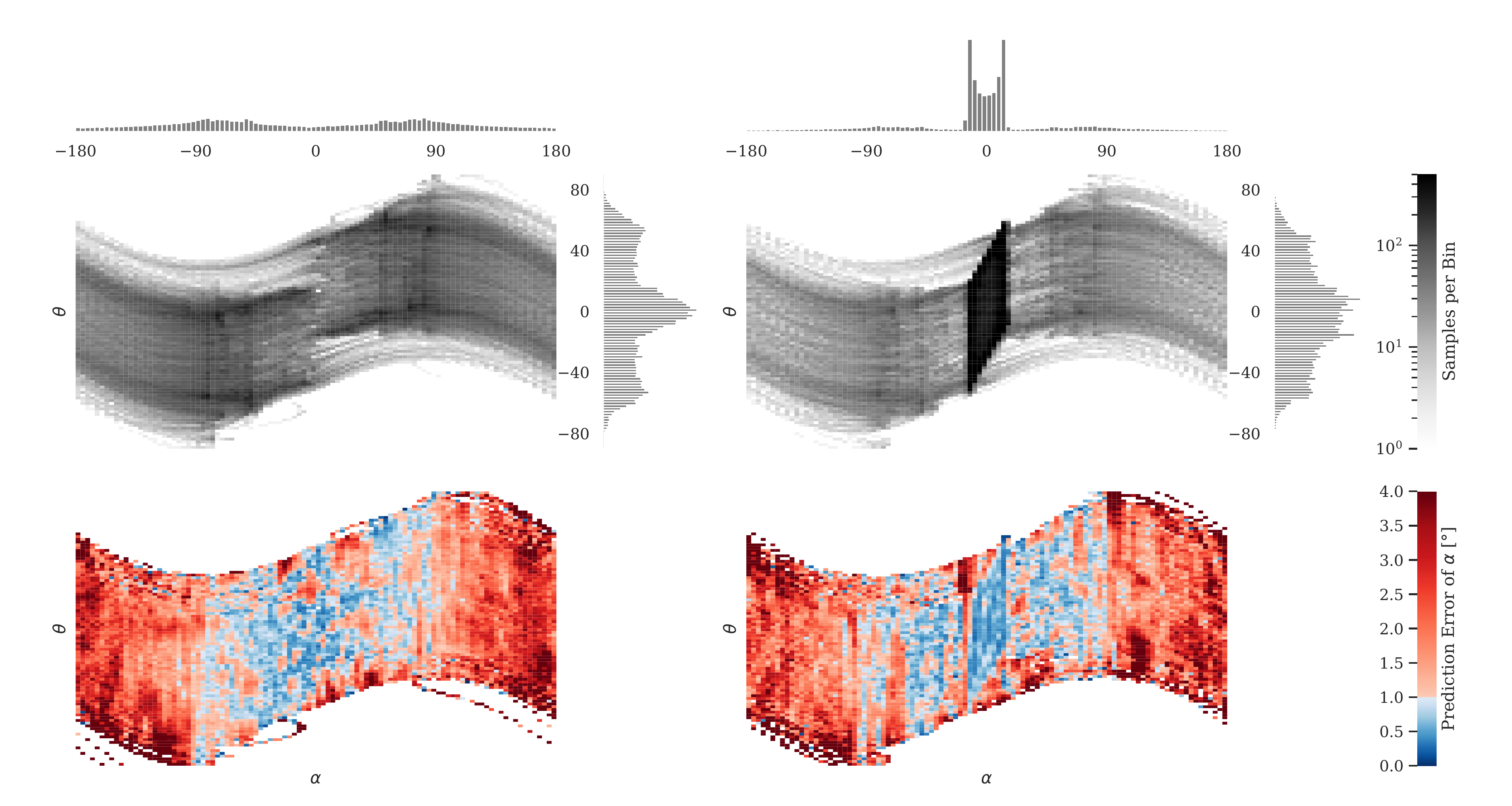}
\caption{Comparison of an unbiased (left) and a data set biased around the upright pendulum position (right), with the absolute data distribution over $\theta$ and $\alpha$ (top), and the resulting prediction error distribution after training of the SE (bottom). Both data sets contain 336'000 images and were trained on the same DNN architecture. While the unbiased data set met the convergence criteria after 48 episodes, the unbiased converged after just four episodes. We also found that the RL agent could adapt much more quickly and more reliably to the SE trained on the biased data set, further reducing the interaction time on the hardware system.}
\label{data-distribution}
\end{figure*}

Despite these measures, the limited network size makes it difficult for the DNN to converge to a low error everywhere. 
Precise state estimates are often not needed throughout the entire state space, and we can evaluate where the SE should be more precise based on the policy analysis conducted in step~\ref{sub:analysis}.
For the Furuta pendulum, we bias the training data set to be more densely sampled around the upper equilibrium point.
An SE trained on a very biased data set can meet our convergence criteria after just four episodes of training. Due to its reliably low prediction error for small angles (compare Figure \ref{data-distribution}), the RL agent could also adapt much faster.\par

To avoid overfitting to the training data set, and to increase the SE's robustness, we also apply data augmentation methods~\cite{perez2017effectiveness} during training.
The input images are randomly zoomed, rotated, shifted, and modulated in brightness (compare Table \ref{hyperparameters-state-estimator}).
While augmenting training data did not increase the accuracy on the validation data set, it substantially increased the prediction performance while testing on the hardware setup.\par

\subsection{Policy Transfer}\label{sub:transfer}
\begin{figure*}
\centering
\includegraphics[width=\linewidth]{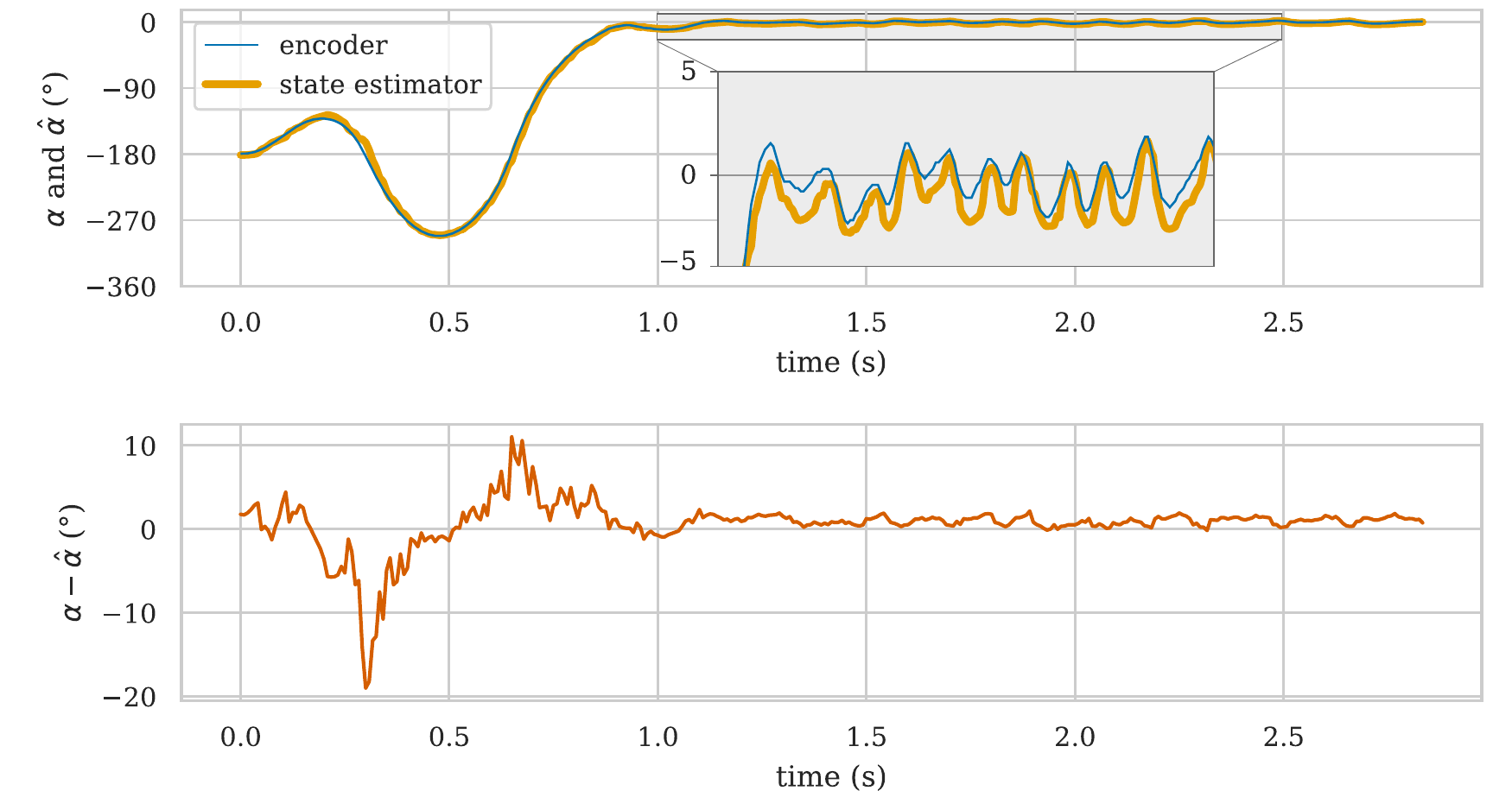}
\caption{Swing-up and balance trajectory of the RL agent with the state estimator as input. Despite much larger errors during swing-up, the policy is able to reliably swing up with only one or two swings, which is faster than a typical run using a standard energy-pumping controller.
}
\label{policy-comparison}
\end{figure*}

Although the state-to-torque policy does not perform well `out of the box' with the DNN-based SE, we found that it can be quickly and easily transferred with additional training, without adjusting any learning hyperparameters.
With an additional training time of approximately \SI{30}{\minute}, the policy adapts to the new input and achieves performance comparable to the policy relying on privileged information. A typical run is shown in Figure~\ref{policy-comparison}, where the pendulum reaches the upright position in a single swing.\par

\section{Discussion}

Our case study demonstrates learning fast and precise pixel-to-torque control of an unstable system, with deep neural networks trained exclusively on real-world data.
Although this task does not pose any theoretical obstacles, the practical challenges are substantial, and to the best of our knowledge, this is the first demonstration on a system requiring a control bandwidth of $\SI{100}{\hertz}$ or higher.
We hope this case study will serve as a starting point for further reproducible and comparable studies. \par
In addition to the usual challenges of using DNNs in control, such as sample efficiency and robustness, we found that achieving both high bandwidth \emph{and} precision becomes a critical challenge for fast and unstable systems. This is especially the case when using function approximators such as neural networks.
While large, sophisticated DNN architectures have tremendous representation power, these networks are not only slower to train but also slower to execute at runtime.
The high bandwidth requirements severely limit the network size, putting speed and precision of the DNN in direct conflict.
In fact, we found that making this trade-off was the dominating factor in designing the learning pipeline.
\par
In order to achieve this while also keeping sample requirements reasonable, we resorted to separating the problems of control and state estimation, and enforcing a state representation based on first principles.
Deep learning literature often advocates against a clear separation~\cite{finn2016spatialAutoencoder, levine2016end} in favor of allowing the learning process to converge to a latent space representation, which may be more parsimonious and task-invariant.
However, in our initial attempts using spatial autoencoders~\cite{finn2016spatialAutoencoder}, we found that a DNN small enough to meet our execution time requirements was by far not expressive enough to learn a meaningful representation. On a data set of more than 400'000 samples, the autoencoder could not detect features precise enough for the RL agent to noticeably increase the reward after an interaction time of over 24 hours.
Learning control and SE separately allowed us to use sample-efficient algorithms, and more easily leverage domain knowledge and privileged information to systematically reduce the DNN size.
Furthermore, for the Furuta system, we can directly compare our learning algorithms against conventional controllers as a clear baseline, which is particularly helpful when debugging unexpected learning outcomes.
\par
One of the advantages we did not initially anticipate is that, by learning the control policy first, we could quantify the robustness of this policy to SE noise.
This provides clear convergence criteria for supervised learning of the SE and is particularly helpful as a quantitative measure for balancing the trade-off in precision and bandwidth.
To cope with limited representation power, we bias the SE training data to regions that require high precision, and compromise in the regions that do not.
With the Furuta pendulum, these regions were simply chosen based on system knowledge, but this is not always straightforward to do for more complex systems.
An alternative we find promising is to not only test the robustness of a standard control policy, but deliberately train robust control policies with a curriculum~\cite{lee2020terrainLocomotion}, for example with progressively noisier environments.
Such a policy would further relax the requirements on the SE, which will be critical for more complex tasks or if even higher bandwidth is required.
To this end, an important direction of research is to quantify the robustness of a control policy~\cite{heim2019learnable}.
\par
In the presented case study, we put minimal effort into making the SE robust to changes in the environment~\cite{laskin2020reinforcementAugment}, and this is certainly an important topic for further studies.
Nonetheless, we found the applied data augmentation methods were crucial to increase the SE's performance on the hardware system. To our surprise, the final policy was quite robust to lighting changes: the LED lamp could be dimmed by \SI{30}{\percent} before the policy failed.
Instead, we believe that the robustness of the learning pipeline itself is an important aspect that is rarely discussed in the literature.
Indeed, we have so far reproduced these results multiple times ourselves, with mixed results.
The entire pipeline was first developed with TensorFlow, then re-implemented in PyTorch essentially unchanged; while learning the SE worked out of the box, the RL agent typically converged to much more aggressive policies, and required additional parameter tuning and testing.
After the pandemic started, the entire hardware setup was moved to Steffen's apartment; here, we were pleasantly surprised that the same pipeline, without any algorithmic changes or hyperparameter tuning, reproduced our results.
However, once the setup was moved back to the lab, it took an unexpected three days of debugging, recollecting training samples, and training from scratch in order to once again reproduce these results.
To better understand how to create reliable learning pipelines, reproducible studies---and reproducing them---are sorely needed.
We believe the vision-based Furuta platform we have presented is ideal for such studies.
Not only does it capture important challenges for fast and unstable systems, it is simple enough that development iterations can be made quickly, and conventional controllers provide not only a clear baseline to compare against, but also a tool to debug and validate different parts of the learning pipeline.
For example, we often determined whether to put more effort into step \emph{A} or step \emph{C} by comparing the DNN-based SE coupled with an LQR controller against the DNN-based policy with encoder readings.
We believe the required effort to recreate the vision-based Furuta platform is reasonable, and we look forward to studies that reproduce, and improve on, the results we have presented.
\section*{Acknowledgment}
\addcontentsline{toc}{section}{Acknowledgment}

We thank Moritz Schneider for his contribution to the code repository for the vision-based Furuta pendulum and his efforts to successfully reproduce the platform and our results. 
We also thank Andreas Doerr, Edgar Klenske, and Niklas Funk for helpful early discussions on the setup of the project.
We thank Tiffany Cheng for help on the video and illustrations.\par
This work was in part funded through the Cyber Valley Initiative.

\printbibliography

\appendix
\added{
\subsection{Training Parameters}
}

\begin{table}[h!]
  \caption{Training Parameters of the PPO Reinforcement Learning Agent}
  \label{rl-hyperparameters}
  \centering
    \begin{tabularx}{\linewidth}{Xc}
      \hline
        \rowcolor{Gray}\textbf{General} & \\
        Interaction time, Training on States & \SI{11}{\hour} \SI{50}{\minute}\\
        (corresponding sample time)  & \SI{8}{\hour} \SI{28}{\minute} \\
        Interaction time, Adaption to SE & \SI{0}{\hour} \SI{28}{\minute} \\
        (corresponding sample time) & \SI{0}{\hour} \SI{16}{\minute} \\
        Sample Frequency & 120 Hz \\
        Horizon & 2048 \\
        Minibatch Size & 32 \\
        Epochs & 10 \\
        Learning Rate & \num{2e-4} \\
        Generalized Advantage Estimation & 0.98 \\
        Discount Factor & 0.995 \\
        Value Function Coefficient & 0.5 \\
        Entropy Coefficient & 0.0 \\
        Clipping & 0.1 \\
    	\rowcolor{Gray}\textbf{Policy Network} & \\
        Type & Multi-Layer-Perceptron \\
        Neurons per Layer & [64, 64, 12] \\
        Activation Function & tanh \\
      \hline
    \end{tabularx}
\end{table}

\begin{table}[h!]
  \caption{Training Parameters of the State Estimator}
  \label{hyperparameters-state-estimator}
  \centering
    \begin{tabularx}{\linewidth}{Xc}
      \hline
        \rowcolor{Gray}\textbf{Data Set} & \\
    	Interaction Time & \SI{28}{\minute} \\
        Sample Frequency & \SI{200}{\hertz} \\
        \rowcolor{Gray}\textbf{General} & \\
        Episodes & 4 (until convergence criteria is met) \\
        Batch Size & 16 \\
        Loss & Mean Squared Error \\
      	Optimizer & ADAM \\
    	\textbf{Data Augmentation} & \\
    	Zoom & (1.0, 1.02) \\
    	x-y-Translation & (-0.01, 0.01) \\
    	Brightness & (0.9, 1.1) \\
      \hline
    \end{tabularx}
\end{table}

\added{
\subsection{Baseline controllers and data generation}\label{appendix-control}

The parameter values are specific to the Quanser Qube Servo 2 pendulum and may need to be adjusted for other Furuta pendulum systems.
The control signal is saturated to stay within motor constraints by simple thresholding. 

\subsubsection{Swing-Up Control}\label{appendix-swing-up}

As a baseline, we use an energy-based control law from~\cite{aastrom2000swinging}.

\begin{equation*}
    u_{\text{swing-up}} = \mu(E_0 - E)\sign(\dot\alpha\cos\alpha)
\end{equation*}

where $\mu$ is a tunable gain, $E_0 = mgl$ is the potential energy of the pendulum at the upright equilibrium, with mass $m$ and length $l$, and $E$ is the current total energy. The $\operatorname{sign}$ operator applies a bang-bang control signal, which results in better performance.
\par
For data collection, the baseline controller is modified to sweep large areas of the state space. We add a PID controller $k_{\text{PID},\theta}$ on the arm angle $\theta$ to follow a reference trajectory $\theta_{\text{ref}}(t) = A_{\text{data},1} \sin(f_{\text{data},1} t)$:

\begin{equation*}
    u_{\text{data,1}} = u_{\text{swing-up}} + k_{\text{PID},\theta}\left(\theta - \theta_{\text{ref}}(t)\right)
\end{equation*}

The PID parameters $k_p$, $k_i$ and $k_d$ and trajectory parameters $A_{data,1}$ and $f_{data,1}$ are listed in Table \ref{data-generation-parameters}.

\subsubsection{Balancing Control}\label{appendix-lqr}

As a baseline, we use a Linear Quadratic Regulator (LQR) controller

\begin{equation*}
    u_{\text{balance}} = -\bm{K}\bm{x}.
\end{equation*}

To design the feedback gain $\bm{K}$ we solve the Ricatti equation of the linear dynamic model of the pendulum at the upright equilibrium $\bm{x}_0 = [0, 0, 0, 0]^T$ and $\bm{u}_0 = 0$. The system matrices of the linear model are provided by Quanser~\cite{quanser2020manual} as

\begin{equation*}
    \bm{A} = 
    \begin{bmatrix}
0 & 0 & 1 & 0 \\
0 & 0 & 0 & 1 \\
0 & 149.3  & -0.01004 & 0 \\
0 & 261.6 & 1 & -0.0103 \\
\end{bmatrix}
\end{equation*}

and $\bm{B} = [0, 0, 49.73, 49.15]^T$.

We found the LQR weight matrices

\begin{equation*}
    \bm{Q} = 
    \begin{bmatrix}
        12 & 0 & 0 & 0 \\
        0 & 5 & 0 & 0 \\
        0 & 0 & 1 & 0 \\
        0 & 0 & 0 & 1 \\
    \end{bmatrix} \\
\end{equation*}

and $R = 1$ to be suitable for reliable and stable data generation trajectories.

To generate data around the upper equilibrium point we perturb the controller with two input signals:

\begin{equation*}
    u_{\text{data,2}} = -\bm{K}\left(\bm{x}-\bm{x_{\text{ref}}}(t)\right)
    + u_{\text{oscillation}}(t)
\end{equation*}

where $u_{\text{oscillation}}(t) = A_{\text{data},1} \sin(f_{\text{data},1} t)$ is a fast oscillation and serves to gather more samples in the region $|\alpha| < \SI{15}{\degree}$, and $\bm{x_{\text{ref}}}(t) = [\theta_{\text{ref}}(t), 0, 0, 0]^T$ is a slow oscillation with $\theta_{\text{ref}}(t) = A_{\text{data},2} \sin(f_{\text{data},2} t)$, and serves to cover large areas of $\theta$. All control and signal parameters are listed in Table \ref{data-generation-parameters}.
\par

\begin{table}[h!]
  \caption{Parameters for Data Generation}
  \label{data-generation-parameters}
  \centering
    \begin{tabularx}{\linewidth}{Xc}
      \hline
        \rowcolor{Gray}\textbf{Energy-based swing-up control} & \\
        $k_p$ & 0.5 \\
        $k_i$ & 0.5 \\
        $k_d$ & 0.05 \\
        $A_{\text{data},0}$ & \SI{60}{\degree} \\
        $f_{\text{data},0}$ & \SI{0.05}{\hertz} \\
        \rowcolor{Gray}\textbf{LQR control for balancing} &  \\
        $A_{\text{data},1}$ & \SI{28}{\volt} \\
        $f_{\text{data},1}$ & \SI{2.4}{\hertz} \\
        $A_{\text{data},2}$ & \SI{30}{\degree} \\
        $f_{\text{data},2}$ & \SI{0.03}{\hertz} \\
      \hline
    \end{tabularx}
\end{table}
}

\end{document}